\documentclass[letterpaper]{article} 
\usepackage{aaai2026}  
\usepackage{times}  
\usepackage{helvet}  
\usepackage{courier}  
\usepackage[hyphens]{url}  
\usepackage{graphicx} 
\usepackage{natbib}  
\usepackage{caption} 
\usepackage{algorithm}
\usepackage{algorithmic}

\usepackage{amsmath} 
\usepackage{amssymb} 

\usepackage{booktabs}

\usepackage{newfloat}
\usepackage{listings}
\DeclareCaptionStyle{ruled}{labelfont=normalfont,labelsep=colon,strut=off} 
\floatstyle{ruled}
\newfloat{listing}{tb}{lst}{}
\floatname{listing}{Listing}
\title{Dual-Phase LLM Reasoning: Self-Evolved Mathematical Frameworks}

\author {
    ShaoZhen Liu\textsuperscript{\rm 1,2},
    Xinting Huang\textsuperscript{\rm 3},
    Houwen Peng\textsuperscript{\rm 3},
    Xin Chen\textsuperscript{\rm 3},
    Xinyang Song\textsuperscript{\rm 1,2},
    Qi Li\textsuperscript{\rm 2},
    Zhenan Sun\textsuperscript{\rm 1,2},
}
\affiliations {
    \textsuperscript{\rm 1} School of Artificial Intelligence, University of Chinese Academy of Sciences\\
    \textsuperscript{\rm 2} Institute of Automation, Chinese Academy of Sciences\\
    \textsuperscript{\rm 3} Tencent Hunyuan Team\\
    liushaozhen2025@ia.ac.cn
}

\usepackage{bibentry}

\begin{document}

\maketitle


\begin{abstract}

  In recent years, large language models (LLMs) have demonstrated significant potential in complex reasoning tasks like mathematical problem-solving. However, existing research predominantly relies on reinforcement learning (RL) frameworks while overlooking supervised fine-tuning (SFT) methods.
  This paper proposes a new two-stage training framework that enhances models' self-correction capabilities through self-generated long chain-of-thought (CoT) data.  During the first stage, a multi-turn dialogue strategy guides the model to generate CoT data incorporating verification, backtracking, subgoal decomposition, and backward reasoning, with predefined rules filtering high-quality samples for supervised fine-tuning. The second stage employs a difficulty-aware rejection sampling mechanism to dynamically optimize data distribution, strengthening the model's ability to handle complex problems.  The approach generates reasoning chains extended over 4× longer while maintaining strong scalability, proving that SFT effectively activates models' intrinsic reasoning capabilities and provides a resource-efficient pathway for complex task optimization. Experimental results demonstrate performance improvements on mathematical benchmarks including GSM8K and MATH500, with the fine-tuned model achieving a substantial improvement on competition-level problems like AIME24. Code will be open-sourced.
\end{abstract}


\begin{figure}[t]
    \centering
    \includegraphics[width=1.0\linewidth]{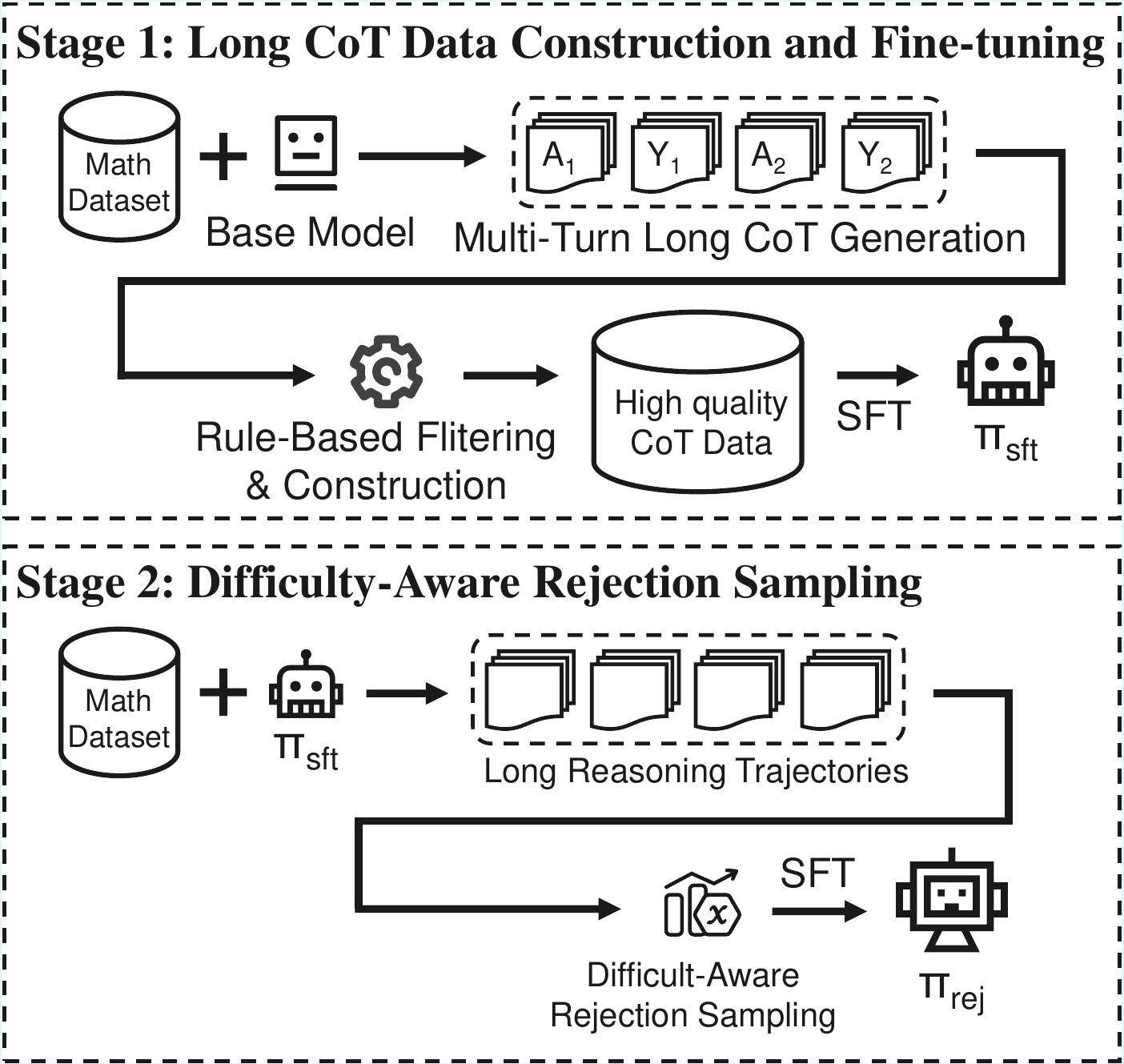}
    \caption{A two-stage self-improvement framework for enhancing LLMs' mathematical reasoning. Stage 1 generates high-quality CoT data via multi-turn reasoning and rule-based filtering for SFT ($\pi_\text{sft}$). Stage 2 employs difficulty-aware rejection sampling on $\pi_\text{sft}$'s outputs to refine reasoning on complex problems, yielding an optimized model $\pi_\text{rej}$}
    \label{fig:overview-of-two-stage}
\end{figure}

\section{Introduction}

Large language models (LLMs) have demonstrated remarkable capabilities in reasoning-related tasks such as mathematics and coding. Notable examples include ChatGPT \cite{OpenAI2023}, Claude \cite{Introducing-2025-07-22}, and Gemini \cite{GeminiTeam2023}. Following the release of GPT-o1 \cite{OpenAI2024} and DeepSeek-r1 \cite{DeepSeek-AI2025}, LLMs with strong reasoning abilities have attracted even more attention, along with inference methods that enhance reasoning.
These models exhibit four capabilities: verification (systematic error-checking), backtracking (abandoning failing approaches), subgoal setting (decomposing problems into manageable steps), and backward chaining (reasoning from desired outcomes to initial inputs) \cite{Gandhi2025}.

The technical details disclosed by DeepSeek-r1 \cite{DeepSeek-AI2025} have revealed a critical step in developing long-chain reasoning models: rule-based reward reinforcement learning. Consequently, current research directions primarily focus on enhancing mathematical and coding performance through reinforcement learning for long reasoning chains \cite{Hu2025,huggingfaceopenr1-2025-07-22,Zeng2025}.

Meanwhile, supervised fine-tuning as a more fundamental approach has often been overlooked. Some paper has already discuss about the relationship between reinforcement learning (RL) and supervised fine-tuning (SFT) \cite{Xiong2025}, but they are mainly based on RL. In this paper, we propose a SFT method for developing long-reasoning models that only requires the model's own generated data for iterative improvement.
Building on these insights, this work introduces a method to enhance LLM reasoning through synthesized long CoT data. Our approach generates multi-step reasoning trajectories that inherently embed four critical capabilities. This data synthesis leverages model-generated content to reduce computational requirements compared to large-model distillation.

\begin{figure}
    \centering
    \includegraphics[width=1.0\linewidth]{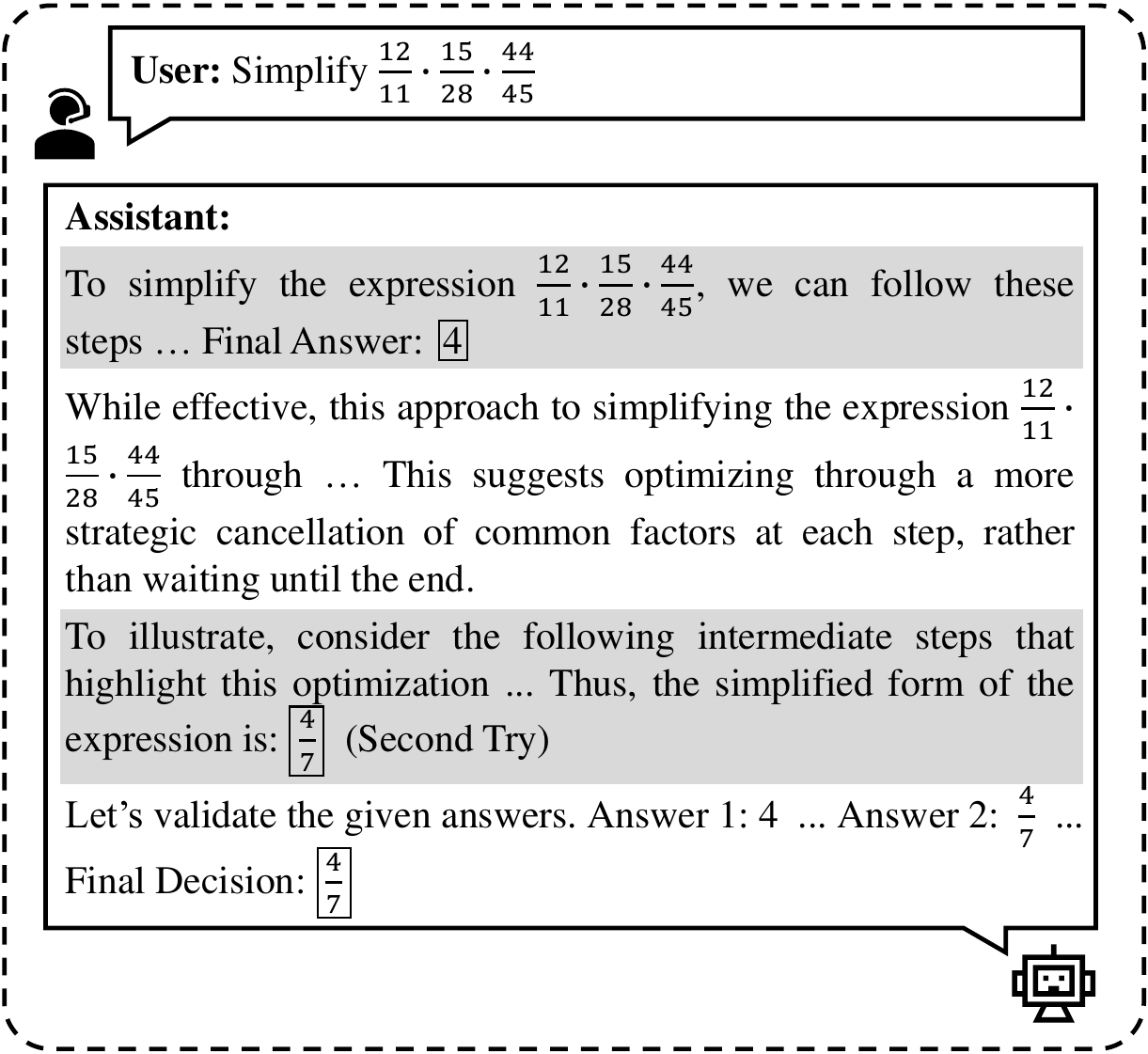}
    \caption{Demonstration of multi-turn self-correction. The model first produces an incorrect answer ``4'' through suboptimal cancellation, then performs backtracking and strategic factorization to identify intermediate optimization opportunities. After recalculating, it verifies correctness and commits to the final refined answer ``$\frac{4}{7}$'', showcasing activation of intrinsic reasoning capabilities.}
    \label{fig:example-output}
\end{figure}

The method employs a new two-stage framework, as illustrated in Figure \ref{fig:overview-of-two-stage}, to generate and utilize self-synthesized long CoT data, relying on the model's own outputs without external dependencies. In the first stage, a multi-turn dialogue strategy guides the model to iteratively generate and refine responses through self-evaluation and correction, as exemplified in Figure \ref{fig:example-output}, which demonstrates how the model identifies errors through backtracking and strategic factorization, then revises its approach to achieve a refined solution. The process repeats over multiple turns to build extended reasoning chains that naturally incorporate all four target capabilities. The generated data is filtered using predefined rules to prioritize high-quality samples, and then used for supervised fine-tuning to enhance the model's intrinsic reasoning abilities. In the second stage, a difficulty-aware rejection sampling mechanism dynamically optimizes data distribution through iterative focus on harder problems, increasing sampling attempts for unsolved questions to collect additional high-quality responses. This refined data is combined with the initial dataset for further fine-tuning, strengthening the model's capacity for complex problem-solving while minimizing computational resource demands.

\begin{itemize}
 \item \textbf{Synthesis Method for Long Chain-of-Thought Data}: 
 Our novel data generation technique produces long reasoning chains containing all four target capabilities with substantially reduced computational overhead compared to large-model distillation. The resulting dataset will be open-sourced.

 \item  \textbf{Supervised Fine-Tuning Framework for Long-Reasoning Models}: 
 We introduce a dual-phase fine-tuning approach where models first train on self-generated reasoning traces, then iteratively improve using filtered high-quality outputs.

 \item \textbf{Empirical Validation}: Experiments demonstrate that models trained with our approach match the performance of those using distilled data from larger models, while exhibiting positive scaling with increased data volume.
\end{itemize}

\section{Related Works}

\subsection{Math Data Synthesis}



Recent advances in LLM mathematical reasoning predominantly rely on GPT-distilled Chain-of-Thought (CoT) data synthesized by frontier models like GPT-4 \cite{yumetamath, wangmathcoder}, as exemplified by NuminaMath \cite{numina_math_datasets} and OpenMathInstruct-2 \cite{toshniwalopenmathinstruct}. While effective, this paradigm inherently limits reasoning capability to the teacher LLM's capacity, as unsolvable problems are excluded from training data.
Although rejection sampling methods \cite{ahn2024large, brown2024large} can improve data quality, this limitation remained unresolved.
We propose difficulty-aware rejection sampling to further address this limitation.

\subsection{Self-improvement Training.}


Self-improvement training currently includes two primary approaches: external reward-based methods \cite{Guan2025, shafayat2025can, zhang2025consistent} using iterative training via Process Reward Models \cite{Lightman2023} or Outcome Reward Models \cite{yang2024qwen2}, and intrinsic reward-based methods that leverage self-generated signals as rewards \cite{taubenfeld2025confidence, kang2025scalable, Yuan2024}. Our approach utilizes self-generated data to foster four reasoning capabilities previously demonstrated to facilitate complex reasoning.


\subsection{LLMs for Supervised Fine-Tuning}

Large-scale supervised fine-tuning (SFT) \cite{radford2018improving, brown2020language, wei2022finetuned, chung2024scaling} constitutes a fundamental approach for enhancing model performance during post-training. SFT adapts pre-trained models to downstream tasks using task-specific datasets, typically formatted as instructions.
Unlike prior works that primarily employ SFT as a foundation for reinforcement learning (RL) \cite{bai2024digirl, zhai2024fine, wu2025reinforcing}, our work strategically utilizes SFT to capture post-RL reasoning patterns. Building on insights from RL outcomes, we develop novel SFT methodologies.

\section{Methodology}



We propose a new multi-turn reasoning strategy that transcends traditional single-turn approaches. This strategy iteratively refines answers through self-verification and backtracking over multiple reasoning turns, enabling systematic error correction. The model synthesizes reasoning history via unified reflection and naturally learns subgoal decomposition for complex problem-solving.

\subsection{Multi-turn Reasoning Strategy}


The multi-turn dialogue strategy operates as follows. After observing the initial prompt $p_1$, the LLM generates its first response:
\begin{equation}
    \textbf{LLM} (p_1) \rightarrow r_1.
    \label{eq:first_response}
\end{equation}
Subsequently, the LLM performs self-evaluation of its analysis process to produce an evaluation text:
\begin{equation}
    \textbf{LLM} (p_1, r_1) \rightarrow e_1.
\end{equation}
Following this analysis, the LLM generates a revised response with further self-correction and evaluation:
\begin{equation}
    \textbf{LLM} (p_2) \rightarrow (r_2, e_2),
\end{equation}
where $p_2 = (p_1, r_1, e_1)$. This iterative process continues for $h$ steps:
\begin{equation}
\begin{aligned}
    &\textbf{LLM} (p_1) \rightarrow r_1, \\
    &\textbf{LLM} (p_1, r_1) \rightarrow e_1, \\
    &\textbf{LLM} (p_2) \rightarrow (r_2, e_2) \quad (p_2 = (p_1, r_1, e_1)), \\
    &\textbf{LLM} (p_3) \rightarrow (r_3, e_3) \quad (p_3 = (p_2, r_2, e_2)), \\
    &\quad\vdots \\
    &\textbf{LLM} (p_h) \rightarrow (r_h, e_h^{\text{final}}) \quad (p_h = (p_{h-1}, r_{h-1}, e_{h-1})).
\end{aligned}
\end{equation}
Here, $p_h$ represents the result after $h$ iterations (with $h$ being predefined), and $e_h^{\text{final}}$ serves as the summary text that analyzes $p_h$ to either accept it as the final answer or select the most reliable answer from previous responses through backward verification.

\subsection{Two-Stage Training Framework}

We propose a two-stage post-training approach for LLMs, which is shown in Figure \ref{fig:overview-of-two-stage}:

\paragraph{Stage 1: Long CoT Data Construction and Fine-tuning.} Starting from a base LLM $\pi$, we collect high-quality multi-turn dialogue data $D_{\text{multi}}$ through our reasoning strategy. This data is used to fine-tune $\pi$ into $\pi_{\text{sft}}$, enhancing four key capabilities: verification, backtracking, subgoal construction, and backward reasoning.

\paragraph{Stage 2: Difficulty-Aware Rejection Sampling Optimization.} We employ improved rejection sampling to collect high-quality responses from $\pi_{\text{sft}}$, filtering them into $D_{\text{rej}}$. Combining $D_{\text{rej}}$ with $D_{\text{multi}}$, we fine-tune $\pi$ again to obtain $\pi_{\text{sft+rej}}$, further boosting model performance.

\subsection{Long CoT Data Construction and Fine-tuning}

After understanding the new multi-turn reasoning strategy and the two-stage post-training framework, the next step is to introduce the first stage of the two-stage framework. 
In this stage, we synthesizes high-quality, long CoT data through multi-turn dialogues. This process systematically captures essential reasoning patterns. The generated data serves as the foundation for supervised fine-tuning, enabling the model to internalize both formal solution structures and underlying problem-solving logics.

\paragraph{Long CoT Data Construction.}

\begin{figure}
    \centering
    \includegraphics[width=1.0\linewidth]{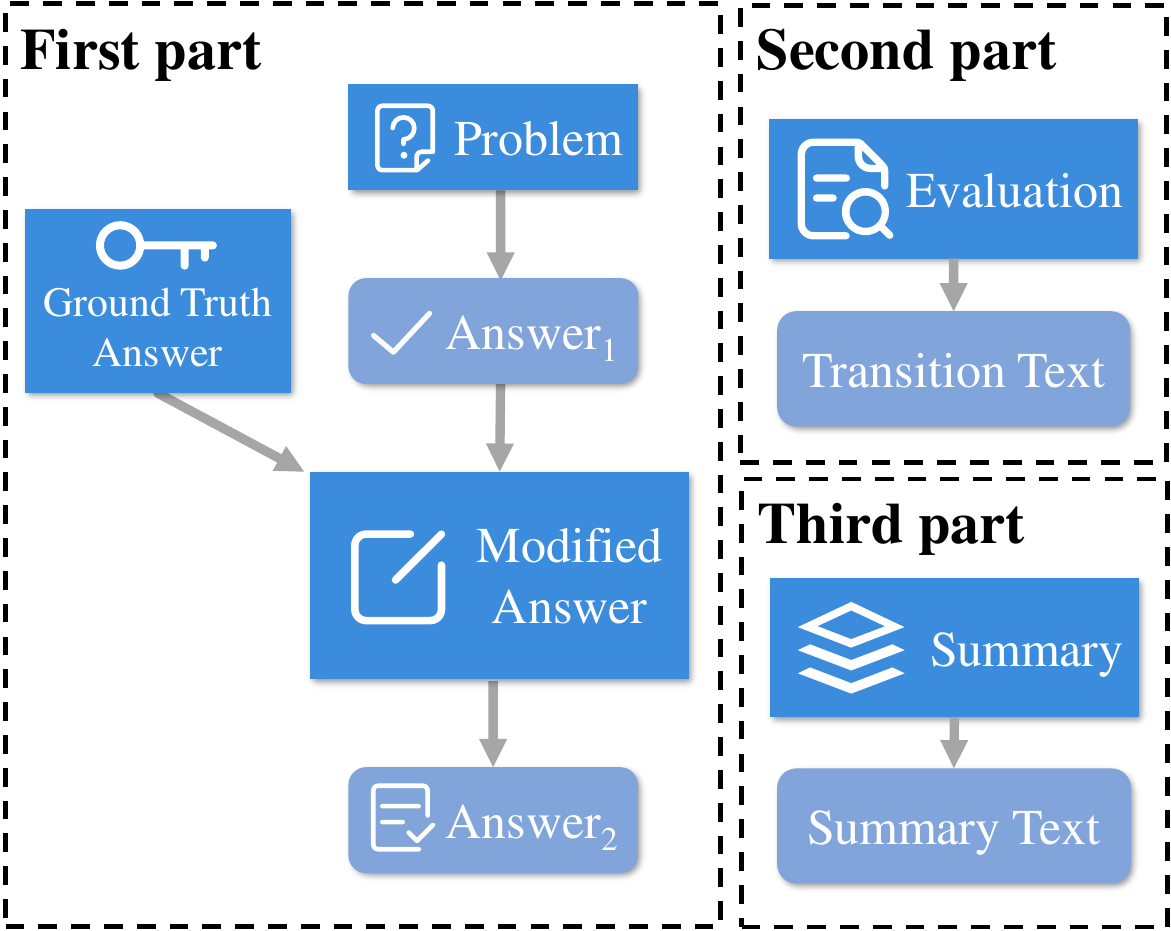}
    \caption{Flow diagram for self-generating Long Chain-of-Thought data. Dual-step answers ($\text{Answer}_1$, $\text{Answer}_2$) are derived from Problem and Ground Truth. Evaluation produces Transition Text through answer analysis; Summary yields synthetic verification.}
    \label{fig:long_cot_data_gen}
\end{figure}

Our long CoT construction strategy is designed based on the multi-turn reasoning strategy introduced earlier, aiming to emulate four logical patterns of mathematical experts: verification, backtracking, subgoal setting, and backward chaining. The long CoT construction involves two main steps: 1) generating candidate texts through multi-turn dialogues with a base model to synthesize long CoT, and 2) refining these responses into high-quality long CoT data using predefined rules.  

\paragraph{Phase 1: Generating Candidate Texts via Multi-Turn Dialogues.}  
Since logical patterns such as verification, backtracking, subgoal setting, and backward chaining rarely emerge spontaneously in base models, single-turn dialogues often fail to capture these reasoning chains. Unlike conventional data synthesis methods, we employ iterative prompting to guide the model to generate distinct reasoning steps independently. These outputs are then combined to form high-quality long CoT data incorporating the four target logical patterns.  


Our approach intentionally diverges from the strict sequential generation sequence $p_1 \rightarrow r_1 \rightarrow e_1 \rightarrow \dots \rightarrow r_h \rightarrow e_h^{\text{final}}$ typical of prior strategies. This adaptation addresses base models' inherent limitations in logical consistency—enforcing rigid sequences often induces context forgetting or instruction deviations.

Instead, we implement a structured two-phase synthesis, visualized in Figure \ref{fig:long_cot_data_gen}, where Problem initiates $\text{Answer}_1$ generation, followed by instruction-triggered $\text{Answer}_2$ production and subsequent Transition Text ($\text{Y}_1$) and Summary ($\text{Y}_2^{\text{final}}$) derivation. For more details about this part, see Appendix.

While theoretically extensible to more turns, strong inter-turn dependencies impede scalable decomposition. Thus, our method strategically limits reasoning to two turns, enabling iterative candidate text synthesis for long CoT construction without external models. This design balances feasibility with the systematic integration of verification and backtracking mechanisms.

\paragraph{Phase 2: Constructing High-Quality Long CoT Data.}  
After obtaining candidate texts, we filter and refine them to build the final dataset \(D_{\text{multi}}\). Each candidate chain \((\text{Problem}, \text{Answer}_1, \text{Y}_1, \text{Answer}_2, \text{Y}_2^{\text{final}})\) is classified into four categories based on the correctness of \(\text{Answer}_1\) and \(\text{Answer}_2\) (denoted as \(a_1\) and \(a_2\)):
\begin{itemize}
    \item True-to-True: Both \(a_1\) and \(a_2\) are correct.  
    \item True-to-False: \(a_1\) is correct, but \(a_2\) is incorrect.  
    \item False-to-True: \(a_1\) is incorrect, but \(a_2\) is correct.  
    \item False-to-False: Both \(a_1\) and \(a_2\) are incorrect.  
\end{itemize}

Among these, True-to-True and False-to-False are common, True-to-False is undesirable, and False-to-True represents the ideal learning target. Following the methodology in this paper \cite{Xiong2025}, our work constructs the fine-tuning dataset with a balanced ratio of \(a_1\)-correct to \(a_1\)-incorrect samples (1:1). Specifically:
\begin{enumerate}
    \item Extract False-to-True data, count the total number \(\text{sum}_1\).
    \item Extract True-to-False data, count the total number \(\text{sum}_2\).
    \item Supplement with True-to-True data to achieve \(\text{sum}_1 - \text{sum}_2\).
\end{enumerate}


Our data construction employs distinct processing strategies for three data types: True-to-True, True-to-False, and False-to-True, rather than uniformly concatenating components as \(\text{Problem} + \text{Answer}_1 + \text{Y}_1 + \text{Answer}_2 + \text{Y}_2^{\text{final}}\).
Additionally, we implemented data filtering concurrently during the concatenation process. Refer to Appendix for implementation details.

It should be noted that the concatenation operator ``$+$'' denotes logical continuation within this context, as opposed to representing simple string concatenation. All filtering thresholds (e.g., 20-character minimum) were determined through experiments conducted on the validation set within the research. The $\boxed{}$ symbol follows mathematical notation conventions for final answer presentation.

\paragraph{Fine-Tuning with Long Chain-of-Thought Data.}

Based on the aforementioned methodology, we generated a large corpus of text data and retained only high-quality samples meeting predefined criteria. We then fine-tuned the model using the long CoT dataset \(D_{\text{multi}}\). The fine-tuning process follows standard practices, aiming to maximize the following objective: 
\begin{equation}
\begin{aligned}
    \mathbb{E}_{x \sim D_{\text{multi}}} & \left[ \log P(Y_1 | x, A_1) \right. \\
    &+ \log P(A_2 | x, A_1, Y_1)\\
    &+ \left. \log P(Y_2^{\text{final}} | x, A_1, Y_1, A_2) \right]. 
\end{aligned}
\label{eq:objective}  
\end{equation}
Notably, prior study \cite{Xiong2025_f4d6} observed that such strategies may induce reward hacking behaviors, where models deliberately generate incorrect initial answers \(A_1\) to artificially improve corrections in \(A_2\), thereby degrading \(A_1\)'s reliability. However, our experiments revealed minimal occurrence of this phenomenon. We attribute this to the balanced data ratio (\(a_1\)-correct : \(a_1\)-incorrect = 1:1), which mitigates skewed optimization incentives.

Detailed configurations of the long CoT fine-tuning process, including hyperparameters and training protocols, are elaborated in Section \ref{section:5.1}.  

\subsection{Rejection Sampling}

Following the previous stage, we obtained a fine-tuned model \(\pi_{\text{sft}}\) capable of generating long CoT outputs using the high-quality dataset \(D_{\text{multi}}\). In this phase, we further optimize the model's performance via an improved rejection sampling algorithm tailored for mathematical reasoning.

Traditional rejection sampling methods typically generate multiple responses per problem and retain only correct solutions.
However, prior studies \cite{NEURIPS2024_0ef1afa0} identified limitations in this approach: it disproportionately favors easier problems, as harder ones are rarely answered correctly within limited trials. For instance, DeepSeekMath-7B achieves 90\% accuracy on MATH500 with 100 samples per problem, indicating that stronger open-source models can generate correct answers for most problems if sufficiently sampled. This motivates our improved strategy to address the underrepresentation of challenging problems.

\paragraph{Difficulty-Aware Rejection Sampling.}  
To prioritize sampling correct responses for difficult problems, we propose an iterative, difficulty-aware rejection sampling algorithm. The workflow for a dataset \(D_0\) is as follows:  
\begin{enumerate}  
  \item For each problem \(S \in D_0\), generate \(n_1\) responses \(x\) using the LLM.  
  \item Retain correct responses and reject incorrect ones.  
  \item Aggregate all retained responses.  
  \item Filter \(D_0\) to create \(D_1\), containing only problems with zero correct responses in \(n_1\) trials.  
  \item Repeat steps 1–4 for \(D_1\) with \(n_2\) samples per problem.  
  \item Iterate this process \(h\) times, progressively focusing on harder problems.  
\end{enumerate}  

This approach ensures harder problems receive exponentially more sampling attempts (e.g., \(n_1 = 2\), \(n_2 = 10\), \(n_3 = 100\) for \(h = 3\)), balancing acceptance rates between easy and hard problems. For retained responses, we select medium-length answers to avoid overly concise (potentially incomplete) or verbose (redundant) reasoning chains. Detailed hyperparameters are provided in Section \ref{section:5.1}.  

\paragraph{Merging Rejection Sampling and Multi-Turn Dialogue Data.}  
The rejection sampling dataset \(D_{\text{rej}}\) is combined with \(D_{\text{multi}}\) to form an augmented dataset \(D_{\text{multi+rej}}\). To distinguish data sources, we append specific markers to each problem:  
\begin{itemize}  
  \item For \(D_{\text{multi}}\): Append \texttt{\textbackslash n\textbackslash nUsing the solution style from multi-turns data.}  
  \item For \(D_{\text{rej}}\): Append \texttt{\textbackslash n\textbackslash nUsing the solution style from rejection data.}  
\end{itemize}  

These markers help the model learn distinct distributions while focusing on their inherent logical patterns. Following previous paper \cite{Zelikman2022}, we fine-tune the base model \(\pi\) on \(D_{\text{multi+rej}}\) using standard protocols. Although fine-tuning \(\pi_{\text{multi}}\) directly on \(D_{\text{rej}}\) showed marginal gains, our merged approach yields superior results. Comparative analyses of alternative fine-tuning strategies (e.g., sequential fine-tuning on \(D_{\text{multi}}\) followed by \(D_{\text{rej}}\)) are detailed in the Experimental Section.

\begin{table*}[t]
\centering
\begin{tabular*}{0.91\linewidth}{l|ccccccc}
\toprule
Method & AIME24 & AMC23 & GSM8K & MATH500 & SVAMP & TabMWP & Gaokao2023en \\
\midrule
Qwen2.5-7B-Instruct & 6.70 & 50.00 & 91.70 & 74.20 & 94.30 & 90.60 & 62.60 \\

+ self-rewarding & 6.70 & 52.50 & 89.50 & 72.00 & 93.00 & 91.20 & 61.30 \\
+ think twice & 10.00 & 50.00 & 90.10 & 73.40 & 93.50 & 90.50 & 62.60 \\
+ $D_{\text{multi}}$ (Ours) & \underline{16.70} & \underline{52.50} & 90.60 & \underline{75.40} & \underline{94.80} & \underline{95.40} & \underline{62.90} \\
+ $D_{\text{multi+rej}}$ (Ours) & 13.30 & 50.00 & 91.10 & 76.00 & 94.10 & 96.30 & 64.40 \\
\quad + multi-turn marker & 13.30 & \textbf{57.50} & \textbf{91.80} & 75.20 & 94.00 & 95.00 & \textbf{64.90} \\
\quad + rejection marker & \textbf{16.70} & 52.50 & 91.70 & \textbf{76.60} & \textbf{95.10} & \textbf{96.20} & 63.40 \\
\midrule
+ distilled & \textbf{26.70} & \textbf{67.50} & \textbf{93.00} & \textbf{84.20} & \textbf{94.40} & \textbf{96.30} & \textbf{75.10} \\
\bottomrule
\end{tabular*}
\caption{Performance Comparison Across Mathematical Reasoning Benchmarks (Accuracy \%)}
\label{tab:main_results}
\end{table*}

\begin{table*}[htbp]
\centering
\begin{tabular*}{0.92\linewidth}{l|ccccccc}
\toprule
Method & AIME24 & AMC23 & GSM8K & MATH500 & SVAMP & TabMWP & Gaokao2023en \\
\midrule
Qwen2.5-7B-Instruct & 1246.33 & 965.87 & 303.91 & 581.53 & 199.34 & 227.49 & 632.43 \\

+ self-rewarding & 2934.47 & 2119.48 & 500.72 & 1262.80 & 360.83 & 319.92 & 1624.54 \\
+ think twice & 1305.32 & 986.58 & 298.37 & 601.51 & 210.37 & 258.40 & 625.48 \\
+ $D_{\text{multi}}$ (Ours) & 4257.62 & 3214.51 & 1141.43 & 2016.70 & 852.12 & 659.14 & 2148.56 \\
+ $D_{\text{multi+rej}}$ (Ours) & 3946.01 & 2114.83 & 378.64 & 1230.42 & 276.38 & 284.28 & 1465.27 \\
\quad + multi-turn marker & 3709.35 & 2395.91 & 444.42 & 1141.43 & 372.35 & 290.77 & 1364.86 \\
\quad + rejection marker & 4004.06 & 2463.98 & 401.69 & 1442.39 & 326.63 & 282.60 & 1211.20 \\
\midrule
+ distilled & 19327.74 & 11962.26 & 2043.77 & 5869.80 & 1693.42 & 1290.40 & 6645.41 \\
\bottomrule
\end{tabular*}
\caption{Response Length Analysis Across Benchmarks (Average Tokens)}
\label{tab:response_length}
\end{table*}

\section{Experiments}

\subsection{Experimental Setup}
\label{section:5.1}
\paragraph{Dataset Selection.}
We employ the publicly available DeepScaleR dataset \cite{DeepScaleR-2025-07-23}, which originates from the DeepScaleR project and contains a substantial collection of annotated mathematical problem-answer pairs. After preliminary inspection and quality-based filtering, we obtain approximately 15k high-quality data points: specifically 10k for long CoT data synthesis, and 5k exclusively for rejection sampling data synthesis. Detailed preprocessing steps are provided in the Appendix.

\paragraph{Model Evaluation.}  
We conduct comprehensive out-of-domain (OOD) assessments across seven benchmarks: AIME24 \cite{numina_math_datasets}, AMC23 \cite{numina_math_datasets}, GSM8K \cite{Cobbe2021}, MATH500 \cite{Lightman2023}, SVAMP \cite{Patel2021}, TabMWP \cite{Lu2022}, and Gaokao2023en. Detailed descriptions of these datasets are provided in Appendix. Our evaluation framework is built upon Qwen2.5-Math's official implementation
, utilizing the vLLM inference engine with generation parameters set to temperature=0.6 and top\_p=1.0. For the more challenging AMC23 and AIME24 datasets, we adopt a robust evaluation strategy of sampling 8 responses per problem, while other benchmarks employ single sampling. A distinctive aspect of our evaluation protocol involves testing three interaction formats: direct problem-solving, multi-turn dialogue with rejection sampling, and multi-turn dialogue with rejection sampling through specialized tagging, ensuring comprehensive analysis of the model's adaptability across diverse scenarios.  

\paragraph{Baselines.}  
We establish rigorous performance baselines through multiple approaches. First, we define the competitive baseline by fine-tuning the Qwen2.5-7B-Instruct \cite{Qwen2024} model with 15,000 high-quality samples distilled from the r1-distilled-32b model \cite{DeepSeek-AI2025} using DeepScaleR \cite{DeepScaleR-2025-07-23}. This distillation process aims to simulate long-chain reasoning patterns while maintaining parameter efficiency. For comparative analysis, we reproduce state-of-the-art methods including self-rewarding \cite{Xiong2025}, implemented through its official codebase, and think twice \cite{Tian2025}, faithfully reconstructed according to the original specifications. Additionally, we integrate results from $D_{\text{multi}}$ fine-tuning into our main experiments for direct comparison. To ensure fairness, all experiments share identical random seeds, with engineering optimizations applied to accelerate think twice's multi-round reasoning in zero-shot settings. 

\begin{table*}[htbp]  
\centering
\begin{tabular*}{0.9\linewidth}{l|ccccccc}
\toprule  
Method & AIME24 & AMC23 & GSM8K & MATH500 & SVAMP & TabMWP & Gaokao2023en \\  
\midrule  
$\pi_{\text{sft}}$ ($D_{\text{multi}}$) & 16.70 & 52.50 & 90.60 & 75.40 & 94.80 & 95.40 & 62.90 \\  
$\pi_{\text{sft}}$ + $D_{\text{rej}}$ & 10.00 & 47.50 & 91.10 & 76.40 & 94.70 & 95.60 & 64.30 \\  
$\pi$ + $D_{\text{rej}}$ & 10.00 & 42.50 & 90.60 & 74.40 & 94.10 & 95.10 & 62.80 \\  
$D_{\text{multi+rej}}$ (w/o marker) & 10.00 & 45.50 & 90.70 & 73.00 & 94.00 & 95.70 & 63.00 \\  
$D_{\text{multi+rej}}$ & 13.30 & 50.00 & 91.10 & 76.00 & 94.10 & \textbf{96.30} & 64.40 \\  
\quad + multi-turn marker & 13.30 & \textbf{57.50} & \textbf{91.80} & 75.20 & 94.00 & 95.00 & \textbf{64.90} \\  
\quad + rejection marker & \textbf{16.70} & 52.50 & 91.70 & \textbf{76.60} & \textbf{95.10} & 96.20 & 63.40 \\  
\bottomrule  
\end{tabular*}  
\caption{Impact of Data Composition in Rejection Sampling (Accuracy \%)}  
\label{tab:rejection_ablation}
\end{table*}

\subsection{Main Results}
\label{section:5.2}

The primary experimental outcomes are presented in Table \ref{tab:main_results}, where all values exhibit a potential rounding error margin of $\pm$0.1\%.
The response length for each experimental result is detailed in Table \ref{tab:response_length}.
The distilled method establishes a soft upper bound for fine-tuning performance (bolded values), consistently outperforming the Qwen2.5-7B-Instruct baseline across all benchmarks. Underlined entries in the $D_{\text{multi}}$ rows indicate superior accuracy compared to both self-rewarding and think twice approaches.
The following analysis is based on the results shown in Tables \ref{tab:main_results} and \ref{tab:response_length}.

\paragraph{Analysis of $D_{\text{multi}}$ Fine-tuning.}
The $D_{\text{multi}}$ approach demonstrates consistent performance improvements over the baseline model across six of seven benchmarks, particularly showing 149\% relative improvement on AIME24 (6.70\% $\rightarrow$ 16.70\%). This enhancement stems from the method's capacity to integrate diverse reasoning patterns during training, enabling better knowledge transfer between symbolic and numerical domains. The cross-task generalization is most evident in MATH500 (74.20\% $\rightarrow$ 75.40\%) and TabMWP (90.60\% $\rightarrow$ 95.40\%), where tabular reasoning benefits from structured data representations. 

Crucially, these performance gains are mirrored in response length dynamics. The $D_{\text{multi}}$ fine-tuning induces substantial increases in token counts across all benchmarks, with averages reaching 4-14× the baseline (e.g., 303.91 $\rightarrow$ 1141.43 on GSM8K). This expansion stems from explicit modeling of diverse reasoning pathways, where complex tabular data (TabMWP: 659.14 vs baseline 227.487) and symbolic proofs (MATH500: 2016.7 vs 581.53) require extended step-by-step explanations. The dramatic growth in AIME24 responses (1246.33 $\rightarrow$ 4257.62 tokens) particularly reflects the method's capacity for generating detailed olympiad-level proofs.

\paragraph{Analysis of $D_{\text{multi+rej}}$ Enhancement.}
The integration of rejection sampling with $D_{\text{multi}}$ ($D_{\text{multi+rej}}$) yields additional gains in complex reasoning tasks. The rejection mechanism particularly enhances MATH500 performance (75.40\% $\rightarrow$ 76.00\%) by filtering low-confidence solutions during training. Test-time marker injection reveals distinct behavioral patterns: rejection markers boost symbolic reasoning (MATH500: 76.60\%), while multi-turn markers improve interactive problem-solving (AMC23: 57.50\%). This dichotomy suggests complementary strengths in different reasoning modalities. 

This accuracy refinement naturally extends to efficiency metrics. Incorporating rejection sampling achieves 21-68\% length reduction compared to $D_{\text{multi}}$ while maintaining competitive accuracy. The rejection mechanism effectively prunes redundant reasoning branches, as evidenced by MATH500 (1230.42 vs $D_{\text{multi}}$'s 2016.7) and Gaokao2023en (1465.27 vs 2148.56). Notably, marker-specific variations reveal task-dependent optimization: multi-turn markers streamline interactive reasoning (AMC23: 2395.91 vs rejection marker's 2463.98), while rejection markers enhance conciseness in symbolic tasks (MATH500: 1442.39 vs multi-turn's 1141.43).

\paragraph{Comparative Analysis with Other Method.}
When compared to self-rewarding and think twice approaches, $D_{\text{multi}}$ variants show superior parameter efficiency and task adaptability. While self-rewarding suffers from reward hacking in formal proofs (GSM8K: 89.50\% vs $D_{\text{multi}}$'s 90.60\%), think twice's iterative reasoning introduces computational overhead without proportional accuracy gains (AMC23: 50.00\% vs $D_{\text{multi+rej}}$'s 57.50\%). The distilled baseline's exceptional performance (Gaokao2023en: 75.10\%) highlights the untapped potential of high-quality synthetic data generation. 

These accuracy advantages are further reinforced when examining response efficiency. The $D_{\text{multi}}$ variants demonstrate superior length-accuracy trade-offs compared to alternatives: While self-rewarding generates 2.6× longer responses than $D_{\text{multi+rej}}$ on GSM8K (500.72 vs 378.64) with lower accuracy (89.50\% vs 91.10\%), think twice's minimal length increase (AMC23: 986.58 vs baseline 965.87) accompanies negligible performance gains. The distilled method's extreme verbosity (AIME24: 19327.74 tokens) highlights $D_{\text{multi+rej}}$'s practical advantage in balancing detail and conciseness - achieving 75-91\% of distilled's accuracy with merely 16-24\% of its response length.

\subsection{Supplementary Experiments}  
\label{section:5.3}

\subsubsection{Study on Using Rejection Sampling Data.}  
\label{subsec:rejection_study}

The experimental matrix examines different strategies for integrating rejection sampling data. Here, the base model $\pi$ is Qwen2.5-7B-Instruct. The results are shown in Table \ref{tab:rejection_ablation}. Row 1 shows the baseline $\pi_{\text{sft}}$ model fine-tuned with $D_{\text{multi}}$, while Row 2 demonstrates subsequent fine-tuning with $D_{\text{rej}}$ data. Row 3 presents direct fine-tuning of the base model $\pi$ using $D_{\text{rej}}$ alone. Row 4 evaluates $D_{\text{multi+rej}}$ without test-time markers, with Rows 5-7 detailing our proposed method variants.

\paragraph{Analysis of Base Model Fine-tuning Strategy.}  
The superior performance of $D_{\text{multi+rej}}$ with markers (Rows 5-7) over the sequential fine-tuning approach (Rows 1-2) reveals critical insights. For instance, as evidenced by AMC23, our method achieves 57.50\% accuracy with multi-turn markers versus 47.50\% in Row 2, indicating a substantial 21\% relative improvement. This outcome suggests that joint training with marker-aware supervision better preserves model plasticity compared to sequential multi-stage fine-tuning.

\paragraph{Analysis of Combined $D_{\text{multi}}$ and $D_{\text{rej}}$ Usage.}  
Comparing analysis between Row 3 ($\pi$+$D_{\text{rej}}$) and Row 4 ($D_{\text{multi+rej}}$ w/o marker) reveals the synergy of combined datasets. The MATH500 accuracy increases from 74.40\% to 76.40\%, demonstrating that integrating $D_{\text{multi}}$'s multi-format reasoning with $D_{\text{rej}}$'s confidence calibration creates complementary learning signals. However, the Gaokao2023en performance drop (62.80\% $\rightarrow$ 63.00\%) suggests domain-specific tuning requirements.

\paragraph{Analysis of Marker-based Data Merging.}  
The marker-enhanced $D_{\text{multi+rej}}$ (Rows 5-7) outperforms markerless integration (Row 4) across all benchmarks, most notably in AIME24 (16.70\% vs 10.00\%). This 67\% relative improvement confirms that explicit marker tokens during training enable better test-time adaptation. The rejection marker's particular effectiveness on MATH500 (76.60\% vs 73.00\%) indicates its role in eliciting structured mathematical proofs.

As a result, our experiments demonstrate that joint fine-tuning with $D_{\text{multi}}$ and $D_{\text{rej}}$ combined with marker-aware training achieves optimal performance. The multi-turn marker configuration excels in interactive scenarios (AMC23: 57.50\%), while rejection markers boost formal proof generation (MATH500: 76.60\%). This validates our hypothesis that explicit task markers during both training and inference phases enable more effective knowledge transfer between reasoning modalities.

\paragraph{Scaling Behaviors.}

The scaling analysis examines performance patterns under progressively varying training data sizes (0.5k-8k samples) ranging from minimal to substantial scales for $D_{\text{multi+rej}}$ fine-tuning, maintaining strictly fixed 1-epoch training duration and consistent hyperparameters across all trials. Evaluation deliberately focuses on two established GSM8K and MATH500 benchmarks. The results are shown in Table \ref{tab:data_scaling} for accuracy and Table \ref{tab:length_scaling} for computational average tokens length.

\paragraph{Accuracy Trend Analysis.}  
Accuracy demonstrates non-linear growth with increasing data volume. The GSM8K performance improves from 91.30\% (0.5k) to 92.30\% (8k), despite temporary dips at intermediate sizes (90.40\% at 1k). MATH500 shows similar progression, peaking at 76.00\% with full 8k data after fluctuating between 73.20-75.20\%. This pattern suggests gradual knowledge integration overcoming initial overfitting tendencies.

\begin{table}[t]  
\centering
\begin{tabular}{l|cc}  
\toprule  
Training Data Size & GSM8K & MATH500 \\  
\midrule  
0.5k & 91.30 & 75.00 \\  
1k & 90.40 & 75.20 \\  
2k & 91.20 & 73.20 \\  
4k & 91.10 & 75.20 \\  
8k & 92.30 & 76.00 \\  
\bottomrule  
\end{tabular}    
\caption{Performance Scaling with Training Data Volume Accuracy (\%)}  
\label{tab:data_scaling}  
\end{table}

\begin{table}[t]  
\centering  
\begin{tabular}{l|cc}  
\toprule  
Training Data Size & GSM8K & MATH500 \\  
\midrule  
0.5k & 348.93 & 894.87 \\  
1k & 11,695.97 & 15,375.14 \\  
2k & 857.27 & 2,687.33 \\  
4k & 819.71 & 1,892.91 \\  
8k & 941.94 & 2,032.73 \\  
\bottomrule  
\end{tabular}  
\caption{Response Length Scaling with Training Data Volume (Average Tokens)}  
\label{tab:length_scaling}  
\end{table}

\paragraph{Response Length Analysis.}  
Average token counts exhibit inverse scaling relationships. Starting from 348.93 tokens (GSM8K) and 894.87 (MATH500) at 0.5k data point, lengths peak dramatically at 11,695.97/15,375.14 tokens with 1k data samples before subsequently stabilizing near the 800-2,000 tokens range band for larger datasets. Significantly, the 8k configuration achieves 941.94/2,032.73 tokens, reflecting improved conciseness despite utilizing 5.3× more training data than 1k.

The method demonstrates positive scaling laws: larger datasets enhance both accuracy (GSM8K: +1.00\%, MATH500: +1.00\% from 0.5k to 8k) and response efficiency (GSM8K length reduction: 11,695.97 $\rightarrow$ 941.94 tokens). This dual improvement confirms the approach's capacity to leverage expanded training data for simultaneous performance gains and output optimization.

\section{Conclusion}

This study demonstrates that SFT with multi-turn dialogue strategies can effectively enhance LLMs' long-chain reasoning capabilities for mathematical tasks. Our approach integrates verification, backtracking, subgoal setting, and backward chaining into synthesized CoT data, enabling base models to achieve performance comparable to distillation-based methods without external models, while reducing computational costs. Key innovations include activating intrinsic reasoning through autonomous error-correction chains and a difficulty-aware rejection sampling mechanism that mitigates bias toward simple problems. Experiments confirm the effectiveness of our approach.

\textit{Limitations.} However, several limitations of our method warrant further investigation. These involve dependence on manual template design and the two-turn constraint's inadequacy for deep multi-step reasoning (e.g., AIME24). Future directions include developing automated template generation through meta-learning, integrating symbolic reasoning engines for rigorous proofs, and creating adaptive sampling algorithms with interpretable difficulty metrics to further advance complex reasoning capabilities while maintaining methodological efficiency.

\bibliography{main}




\makeatletter
\@ifundefined{isChecklistMainFile}{
  \newif\ifreproStandalone
  \reproStandalonetrue
}{
  \newif\ifreproStandalone
  \reproStandalonefalse
}
\makeatother

\end{document}